\documentclass{article}

\usepackage[final,nonatbib]{nips_2018}

\usepackage[utf8]{inputenc} 
\usepackage[T1]{fontenc}    
\usepackage{hyperref}       
\usepackage{url}            
\usepackage{booktabs}       
\usepackage{amsfonts}       
\usepackage{nicefrac}       
\usepackage{microtype}      

\usepackage{graphicx}
\usepackage{cite}
\usepackage{threeparttable}

\title{Compensated Integrated Gradients to Reliably Interpret EEG Classification}

\author{
  Kazuki Tachikawa, Yuji Kawai,  Jihoon Park, Minoru Asada\\
  Graduate School of Engineering,\\
  Osaka University\\
  \texttt{kazuki.tachikawa@ams.eng.osaka-u.ac.jp} \\
}

\begin{document}
\maketitle
\begin{abstract}
    Integrated gradients are widely employed to evaluate the contribution of input features in classification models because it satisfies the axioms for attribution of prediction.
    This method, however, requires an appropriate baseline for reliable determination of the contributions.
    We propose a compensated integrated gradients method that does not require a baseline.
    In fact, the method compensates the attributions calculated by integrated gradients at an arbitrary baseline using Shapley sampling.
    We prove that the method retrieves reliable attributions if the processes of input features in a classifier are mutually independent, and they are identical like shared weights in convolutional neural networks.
    Using three electroencephalogram datasets, we experimentally demonstrate that the attributions of the proposed method are more reliable than those of the original integrated gradients, and its computational complexity is much lower than that of Shapley sampling.
\end{abstract}

\section{Introduction}
Deep learning has become a promising method for image recognition, natural language processing, speech recognition, and even classification of diseases \cite{thodoroff2016learning,tachikawa2018effectively} and evaluation of brain and body physiology \cite{tsinalis2016automatic,schirrmeister2017deep} from raw electroencephalogram (EEG) signals.
In EEG classification, besides the results it is important to determine the reasons of the classification for its proper interpretation by a specialist.
For instance, in medical decision support systems, interpretation enables doctors to corroborate automatic classification based on machine learning against medical knowledge and possibly unveil previously unnoticed features of a disease.

Several methods for visualizing the separate contribution of input features to classification have been proposed \cite{sundararajan2017axiomatic,schirrmeister2017deep,li2017targeting,
lundberg2017unified,shrikumar2017learning}.
Among them, integrated gradients (IG) and Shapley sampling (SS) have been shown to be theoretically superior to other methods, because they satisfy axioms for the fair attribution of contributions \cite{sundararajan2017axiomatic,vstrumbelj2014explaining,lundberg2017unified}.
Although the IG method is computationally efficient, it requires an appropriate baseline (reference point) for determining reliable contributions.
The baseline is an input assumed to not include any features and is empirically set (usually, the zero point), as no formal methods for finding the appropriate baseline have been devised.
Setting an inappropriate baseline can undermine the reliability of the attributions \cite{2017arXiv171100867K}.
In contrast, the SS method does not require setting a baseline \cite{vstrumbelj2014explaining},
but its computational cost is extremely high \cite{lundberg2017unified}.

In this study, we propose a method for compensating the contributions obtained from IG at an arbitrary baseline by using SS contributions.
The proposed method satisfies the same axioms as the IG method for an appropriate baseline under specific classifier constraints.
We experimentally evaluate the reliability and computational complexity of the proposed method on three EEG datasets.

\section{Compensated IG}
\begin{figure}[t]
    \begin{center}
        \includegraphics[width=13cm,clip]{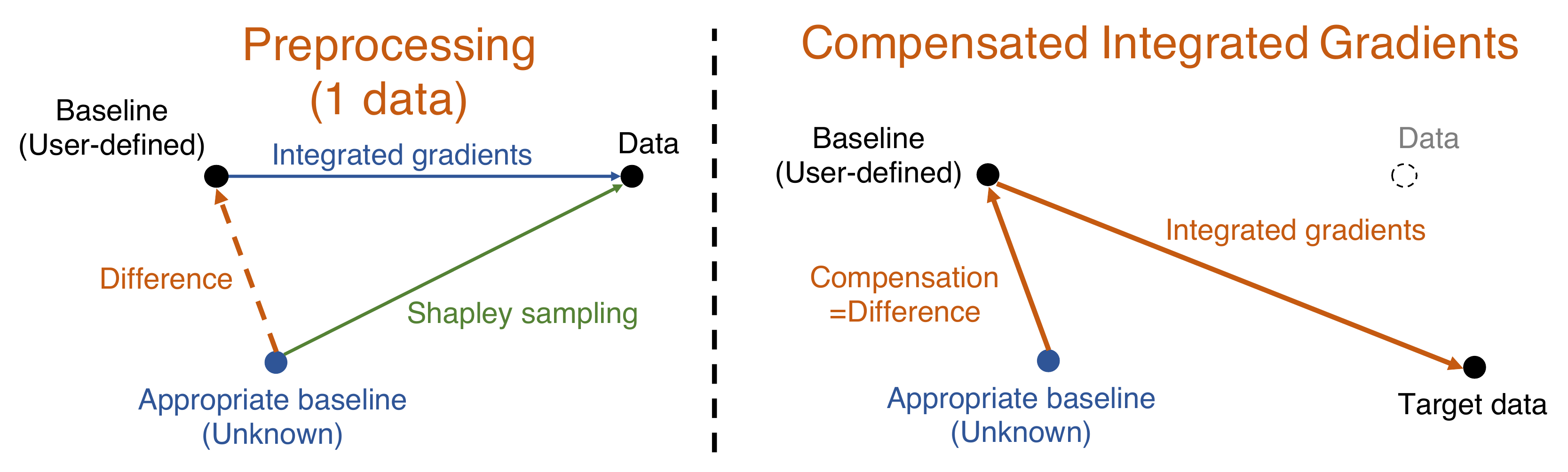}
        \caption{Diagram of the proposed method.
        Left panel: process to obtain the compensation amount.
        Right panel: path to compute a reliable contribution.
        \label{fig:diagram}}
    \end{center}
\end{figure}
The proposed compensated IG method is a type of path method \cite{friedman2004paths,sundararajan2017axiomatic}
that integrates gradients of the output with respect to the input of a classifier, usually before a softmax function in the output layer, along an arbitrary path from the baseline to the input data point.
Given path function $\gamma (\alpha)$ for $\alpha \in [0, 1]$, classifier $f$, and input data $x$, the contribution of the $i$-th feature, ${\rm PathIG}_i^\gamma (x)$, is given by
\begin{eqnarray*}
    {\rm PathIG}_i^\gamma(x) = \int_{\alpha=0}^1 \frac{\partial f(\gamma(\alpha))}{\partial\gamma_i(\alpha)} \frac{\partial\gamma_i(\alpha)}{\partial\alpha} d\alpha .
\end{eqnarray*}
This method satisfies the following four axioms \cite{friedman2004paths,sundararajan2017axiomatic}:
\begin{enumerate}
    \item Implementation invariance: If two classifiers are functionally equivalent, i.e., they retrieve equal outputs for every input, the attribution of contributions are also equivalent.
    \item Dummy: If a classifier does not depend on some variables, the contribution of these variables is always zero.
    \item Linearity: Let classifier $f$ be represented by the weighted linear sum of two sub-classifiers $f_1$ and $f_2$, i.e., $f=af_1+bf_2$ for scalars $a$ and $b$.
    Then, the contribution of the classifier $\phi(f)$ is also obtained by the weighted linear sum of the contribution of each sub-classifier:
    $\phi(f)=a\phi(f_1)+b\phi(f_2)$.
    \item Completeness: The sum of the contributions of all features equals the output value of the classifier.
\end{enumerate}
If the path is a straight line, the path method further satisfies the symmetry axiom defined below (this method is called IG) \cite{sundararajan2017axiomatic}.
\begin{enumerate}
    \setcounter{enumi}{4}
    \item Symmetry: Let the output of a classifier not always change, even if exchanging features $x_i$ and $x_j$, i.e., $f(x_i, x_j)=f(x_j, x_i)$. Then, the contributions of these features are identical.
\end{enumerate}
The SS method also satisfies all of these axioms, and therefore, this method corresponds to the IG method
appropriately setting the reference point (green arrow in the left panel of Fig. \ref{fig:diagram}).

In most cases, a zero input is used as a baseline instead of a better input (blue arrow in the left panel of Fig. \ref{fig:diagram}).
However, this zero input is often inappropriate, resulting in unreliable attribution of prediction \cite{2017arXiv171100867K}.
To compensate the contribution from the user-defined zero input, we calculate the difference between the contribution obtained from this input and a reliable contribution obtained using the SS method for one data record (orange dashed arrow in the left panel of Fig. \ref{fig:diagram}).
Note that the computational cost of this step is not very high because the SS method is only applied to one data record.
This difference corresponds to the integral of the gradients along an unknown path from the true baseline to the user-defined baseline.
The value of the integral does not depend on the data, as it is determined only by the two baselines.
The integral value is added to the contribution obtained from the user-defined baseline and into an arbitrary target data point to obtain the input/output gradients integrated along the orange path in the right panel of Fig. \ref{fig:diagram}.

This operation satisfies axioms 1-4, besides the symmetry axiom (5) if the classifier processes of the input features are independent and identical, as proved in Appendix \ref{appendix:proof}.
For example, the identity constraint is realized by a weight sharing technique in convolutional neural networks (CNNs).
However, the contributions in spatial CNNs, which are widely used in image recognition, do not satisfy the symmetry axiom because each filter processes several input features, and therefore, they violate the independence constraint.
In contrast, when using temporal CNNs, in which each filter convolutes an input time-series feature, the contributions satisfy all the axioms.
Still, an inappropriate baseline in the original IG produces the violation of all axioms.
\begin{table*}[t]
    \begin{center}
        \caption{Spearman's correlation compared to contributions obtained from Shapley sampling. The temporal model corresponds to one-dimensional convolutional neural networks (CNNs), and the spatiotemporal model to two-dimensional CNNs.
        \label{table:result}}
        \begin{tabular}{ccccc|ccccc}\hline
            \multicolumn{5}{c|}{Temporal Model} &  \multicolumn{5}{c}{Spatiotemporal Model}  \\
            Dataset & Class & C-IG & SS & IG & Dataset & Class & C-IG & SS & IG \\\hline
            PhysioNet & N1 & \bf0.983  & 0.970  & 0.180  &  &  & &  &  \\
             & N2 & \bf0.970  & 0.953  & 0.655  &  &  &  &  &  \\
             & N3 & \bf0.988  & \bf0.988  & 0.925  &  &  &  &  &  \\
             & R & 0.963  & \bf0.970  & 0.665  &  &  &  &  &  \\
             & W & \bf0.987  & 0.972  & 0.326  &  &  &  &  &  \\
            CHB-MIT & Szr. & \bf0.996  & 0.994  & 0.817  & CHB-MIT & Szr. & 0.806  & \bf0.983  & 0.695  \\
             & No S. & \bf0.993  & 0.990  & 0.293  &  & No S. & 0.917  & \bf0.982  & -0.037  \\
            UCI EEG & Alc. & \bf0.994  & 0.991  & 0.260  & UCI EEG & Alc. & 0.793  & \bf0.989  & 0.323  \\
             & Ctr. & \bf0.995  & 0.992  & 0.258  &  & Ctr. & 0.739  & \bf0.988  & 0.331  \\\hline
        \end{tabular}
        {\small
            C-IG, compensated integrated gradients (proposed method);
            SS, Shapley sampling;
            IG, integrated gradients; \\
            R, rapid eye movement;
            W, wakefulness;
            Szr., seizure;
            No S., no seizure;
            Alc., alcoholism;
            Ctr., control.
        }
    \end{center}
\end{table*}

\section{Experiments}
We evaluated the reliability and computational cost of the proposed method on three publicly available EEG datasets, namely, the PhysioNet polysomnography dataset  \cite{goldberger2000physiobank,kemp2000analysis,PhysioNet},
UCI EEG dataset \cite{zhang1995event,UCI},
and CHB-MIT Scalp EEG dataset \cite{shoeb2009application,CHBMIT}.
For the PhysioNet polysomnography dataset, we trained six-layer temporal CNNs to classify the data into five sleep stages (N1, N2, N3, rapid eye movement, and wakefulness).
These CNNs are one-dimensional and convolve input EEG signals separately from each other to acquire time-domain features that are integrated in the fully connected output layer.
For both the UCI EEG and CHB-MIT Scalp EEG datasets, we trained five-layer temporal CNNs and four-layer spatiotemporal CNNs to classify the data into two classes (alcoholism/control and seizure/no seizure, respectively).
The spatiotemporal CNNs are two-dimensional and prevent the proposed method from satisfying the symmetry axiom.

In each dataset, we randomly selected 200 data records of each class.
For each classifier, we computed the contributions of input features using the proposed method, the IG method using the zero-input baseline, and the SS method.
We used 10 additional data records on the proposed method for compensation to mitigate the sampling error in the SS method, although theoretically, one data record should be enough for compensation.
Ideally, contribution comparison should be performed against true contributions, but they are unknown.
Therefore, we considered the contributions obtained by the complete SS method as the true contributions and measured the similarity among contributions by the Spearman's correlation \cite{ghorbaniinterpretation,adebayo2018local}.
Large correlation coefficients (close to 1) indicate high similarity between two compared contributions.
Note that the comparison with the contributions obtained by the SS method reflects the sampling errors.
Furthermore, we determined the contributions of EEG sensors (electrodes) to the classification of alcoholism in the UCI EEG dataset to qualitatively compare the methods.
We used a temporal model and averaged the contributions from the 200 analyzed data records for this comparison.

\section{Results}
The Spearman's correlation coefficients of all datasets and models are listed in Table \ref{table:result}.
The coefficient values of the proposed method, the compensated IG, were almost the same as those of SS on the temporal CNNs.
In contrast, IG exhibited the lowest values, particularly in the UCI EEG dataset.
Therefore, the proposed method can suitably compensate the unreliable contributions obtained from the IG method.

Even when the compensated IG does not satisfy the symmetry axiom on the spatiotemporal CNNs, its coefficient values are larger than those of the original IG with its zero-input baseline.
However, the values of the compensated IG were lower than those of SS in this case, indicating that the violation of the symmetry axiom undermines the reliability of the estimated contributions.
Still, compensation improved the reliability of the original IG.

We defined the computational complexity by the number of backpropagations (for IG) $\times$ number of data records $+$ number of forward propagations (for SS) $\times$ number of sensors (electrodes) $\times$ number of data records.
Assuming equal computational costs for forward and backpropagation, we obtained computational costs of 40,000 for IG, 650,000 for the proposed method, and 12,200,000 for SS when applied to the UCI EEG dataset.
If 1000 data points are targeted, the computational cost ratio is 20:81:6100.

\begin{figure*}[t]
    \begin{center}
        \includegraphics[width=13cm,clip]{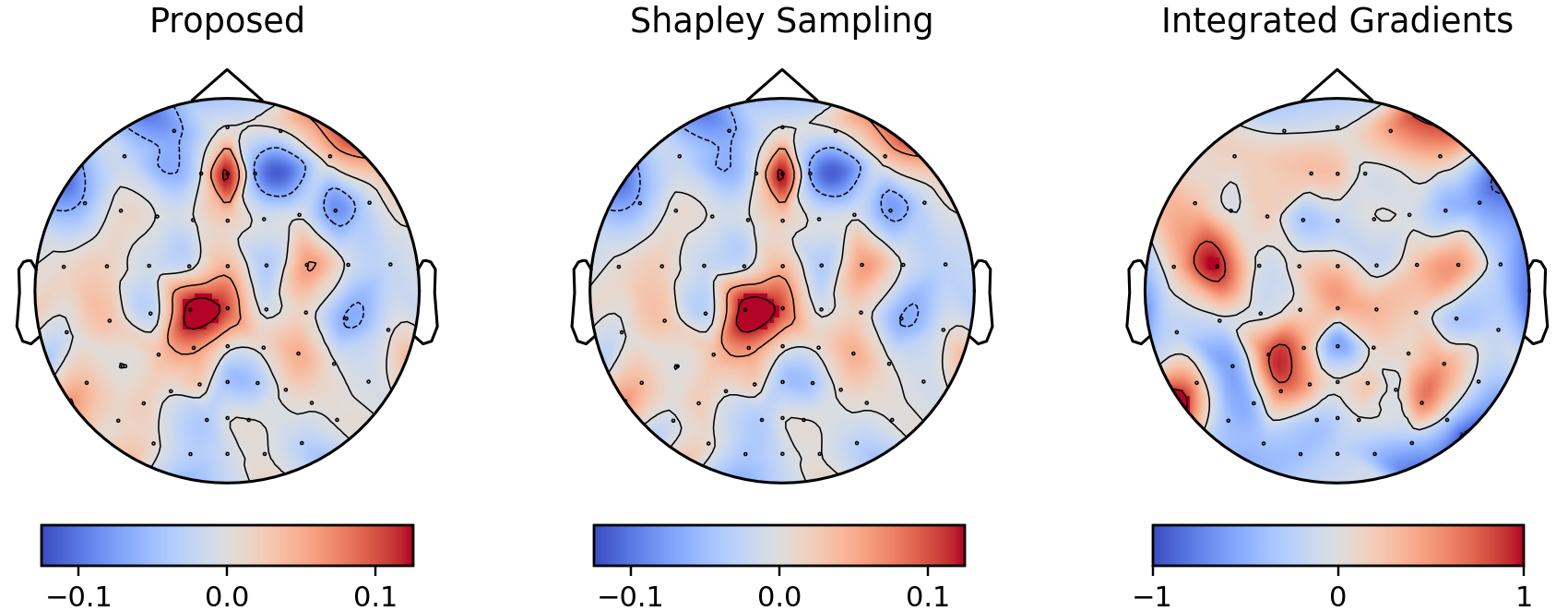}
        \caption{Contributions of EEG sensors (electrodes) over the scalp to classify alcoholism.
        The average contribution of the methods applied to 200 data records from the UCI EEG dataset is depicted.
        The red and blue areas represent positive and negative contributions to the classification of alcoholism, respectively.
        \label{fig:scalpPlot}}
    \end{center}
\end{figure*}

Fig. \ref{fig:scalpPlot} shows the contributions of the EEG sensors (electrodes) for alcoholism classification.
The contributions obtained from the proposed method (left panel) were indistinguishable from the true contributions obtained from the SS method (middle panel), whereas those obtained from the original IG method (right panel) exhibit a different distribution from those obtained from the other methods.

\section{Conclusion}
We propose a compensated IG method using SS to improve the reliability to interpret classification outcomes.
The proposed method satisfies four axioms, namely, implementation invariance, dummy, linearity, and completeness, besides an additional symmetry axiom under classifier constraints (see Appendix \ref{appendix:proof}).
Using three EEG datasets, we demonstrate that the proposed method can compute more reliable contributions than the IG method with an inappropriate baseline and presents much lower computational cost than the SS method.
The contributions obtained from the proposed method were very similar to those obtained from the SS method especially for temporal CNNs, which meet the constraints for the symmetry axiom.
However, the classifier constraints decrease the classification accuracy (see Appendix \ref{appendix:acc}).
In contrast, spatiotemporal CNNs exhibit higher classification accuracy but lower interpretation reliability than the temporal CNNs.
Therefore, classifier selection should depend on whether reliability or classification accuracy are emphasized.
Even when the symmetry axiom is violated given the convolution among input features, we demonstrated that compensation effectively improves the reliability of the original IG.
In future developments, we will apply the proposed method to multichannel time-series data different from EEG, e.g., acceleration of human activities \cite{zeng2014convolutional} and electrocardiography \cite{zheng2014time}.

\subsubsection*{Acknowledgments}
This work was supported by the Center of Innovation Program from Japan Science and
Technology Agency and JST CREST Grant Number JPMJCR17A4, Japan.

\bibliographystyle{unsrt}
\bibliography{myrefs}

\newpage
\appendix
\section*{Appendices}

\section{Proof: all path methods satisfy the symmetry axiom when using temporal CNNs \label{appendix:proof}}
Let an input time-series be $n$-dimensional: $x = [x_1, ..., x_i, ..., x_n]$,
the length of each time-series $z$ be $L$: $ x_i = [z_{i1}, ..., z_{ik} ..., z_{iL}] $, and
a classifier $f(x)$ be given by
\begin{equation}\label{eq:model}
  f(x) = \sum_{i}^{n} \sum_{j}^{m} W_{ij} g^m(x_i),
\end{equation}
where, $W_{ij}$ denotes a weight, $g^m(x_i)$ is a nonlinear function that converts time-series input $x_i$ into outputs $m$.
For the CNN considered in this study, $g^m(x_i)$ represents one-dimensional (temporal) convolutional layers or filters that share weights and $ W_{ij}$ represents the fully connected layer.
The contribution of $p$-th feature $x_p$ is represented as $\phi(f, x_p)$.

From Eq. (\ref{eq:model}),
\begin{eqnarray*}
    \phi(f, x_p)=\phi \left(\sum_{i}^{n} \sum_{j}^{m} W_{ij} g^m(x_i), x_p\right).
\end{eqnarray*}
Using linearity,
\begin{eqnarray*}
    \phi \left(\sum_{i}^{n} \sum_{j}^{m} W_{ij} g^m(x_i), x_p\right)=\phi \left(\sum_{j}^{m} W_{pj} g^m(x_p), x_p\right).
\end{eqnarray*}
Using the implementation invariance axiom and the symmetry assumption (let symmetric features be $ x_q $),
\begin{eqnarray*}
    W_{pj}=W_{qj}.
\end{eqnarray*}
and
\begin{eqnarray*}
    g^m(x_p)=g^m(x_q).
\end{eqnarray*}
Therefore,
\begin{eqnarray*}
    \phi \left(\sum_{j}^{m} W_{pj} g^m(x_p), x_p\right)=\phi \left(\sum_{j}^{m} W_{qj} g^m(x_q), x_q\right).
\end{eqnarray*}
Then,
\begin{eqnarray*}
    \phi(f, x_p)=\phi(f, x_q).
\end{eqnarray*}
Consequently, all path methods satisfy the symmetry axiom if the processes of input features are independent and identical in a classifier.

\section{Classification Accuracy\label{appendix:acc}}

\begin{table*}[h]
    \begin{center}
        \caption{
            Classification accuracy of different models on three datasets.
        }
        \begin{tabular}[t]{ccc}\hline
            Dataset & Model & Accuracy \\\hline
            PhysioNet & Temporal & 81.6\% \\
            CHB-MIT & Temporal & 83.4\% \\
             & Spatiotemporal & 88.7\% \\
            UCI EEG & Temporal & 69.5\% \\
             & Spatiotemporal & 77.1\% \\\hline
        \end{tabular}
    \end{center}
\end{table*}

\end{document}